\documentclass[conference]{IEEEtran}
\IEEEoverridecommandlockouts
\usepackage{cite}
\usepackage{amsmath,amssymb,amsfonts}
\usepackage{algorithmic}
\usepackage{graphicx}
\usepackage{textcomp}
\usepackage{xcolor}
\usepackage{overpic}
\usepackage{hyperref}
\usepackage{pifont}
\usepackage{amsmath,amssymb}
\usepackage{caption}
\usepackage{subcaption}

\DeclareMathOperator{\EX}{\mathbb{E}}

\newcommand{\eg}{\textit{e.g.}}
\newcommand{\ie}{\textit{i.e.}}

\begin{document}

\title{Exploiting Semantics in Adversarial Training for Image-Level Domain Adaptation}

\author{\IEEEauthorblockN{1\textsuperscript{st} Pierluigi Zama Ramirez}
\IEEEauthorblockA{\textit{University of Bologna} \\
pierluigi.zama@unibo.it}
\and
\IEEEauthorblockN{2\textsuperscript{nd} Alessio Tonioni}
\IEEEauthorblockA{\textit{University of Bologna} \\
alessio.tonioni@unibo.it}
\and
\IEEEauthorblockN{3\textsuperscript{rd} Luigi Di Stefano}
\IEEEauthorblockA{\textit{University of Bologna} \\
luigi.distefano@unibo.it}
}

\maketitle

\begin{abstract}
Performance achievable by modern deep learning approaches are directly related to the amount of data used at training time. Unfortunately, the annotation process is notoriously tedious and expensive, especially for pixel-wise tasks like semantic segmentation. Recent works have proposed to rely on synthetically generated imagery to ease the training set creation. However,  models trained on these kind of data usually under-perform on real images due to the well known issue of domain shift. We address this problem by learning a domain-to-domain image translation GAN to shrink the gap between real and synthetic images. Peculiarly to our method, we introduce semantic constraints into the generation process to both avoid  artifacts and guide the synthesis. To prove the effectiveness of our proposal, we show how a semantic segmentation CNN trained on images from the synthetic GTA dataset adapted by our method can improve performance by more than 16\% mIoU with respect to the same model trained on synthetic images.

\end{abstract}


\section{Introduction}
\label{sec:intro}

Recent advancements in computer vision are characterized by a widespread adoption of deep learning, either as end-to-end complete solutions or as components of more complex pipelines. 
A common trait across all the different flavours of deep learning models is the strong correlation between the size of the accurately annotated training set and the achievable performance. As for research work, the need of a large corpus of   annotated data may not be an issue thanks to availability of many curated dataset. Yet, the data issue is limiting a more widespread adoption of deep learning in many practical applications. Even assuming availability of training images for the target environment, manually producing the annotations is a tedious and expensive operation, that quickly become hard to scale for more complex tasks. For example, annotating a single image with a global label usually requires few seconds while annotating the same image for pixel-wise prediction tasks, such as depth estimation or semantic segmentation, requires many minutes or hours, even with the help of professional tools\cite{Cordts_2016_CVPR}. 


\begin{figure}
    \centering
    \includegraphics[width=0.9\columnwidth]{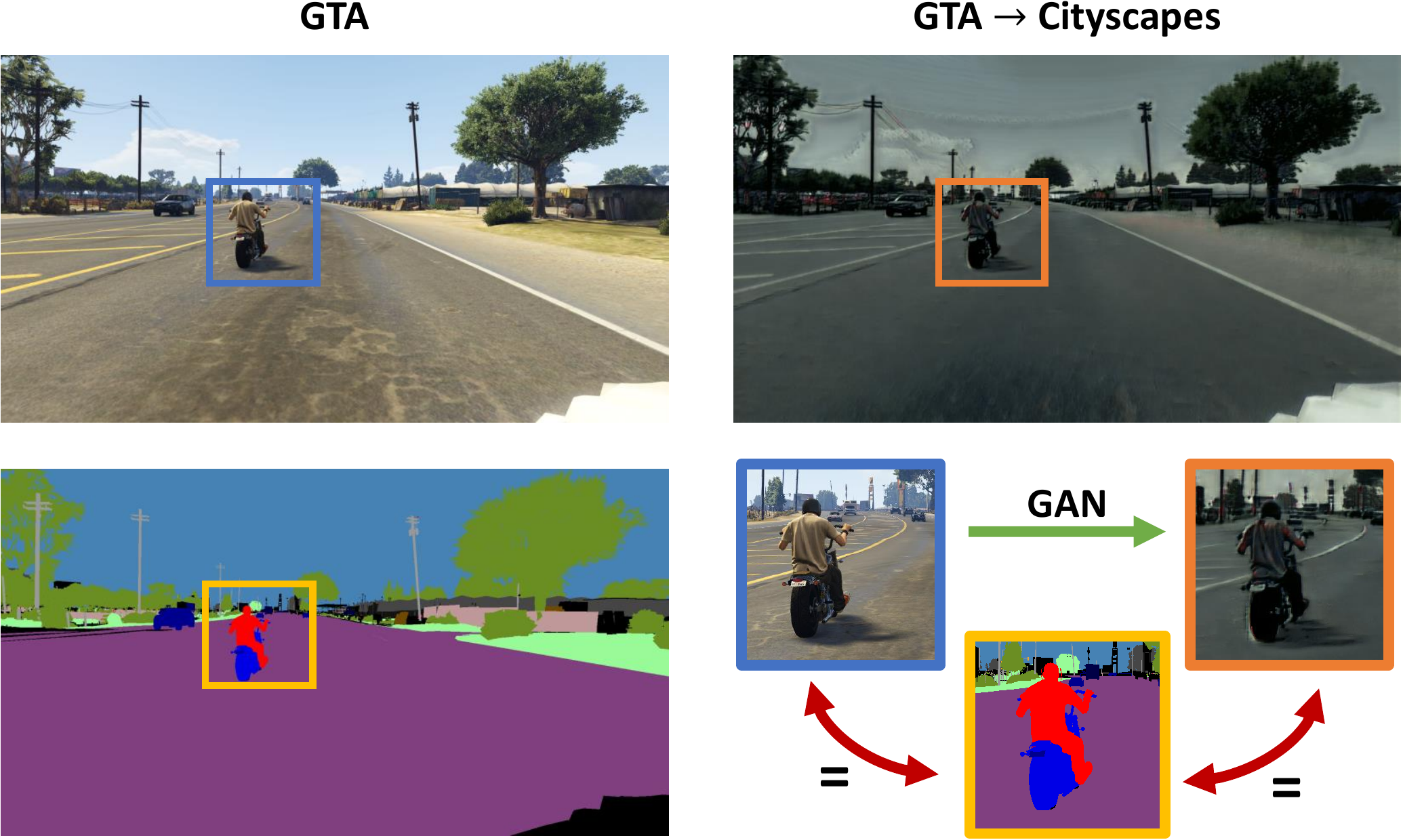}
    \caption{On the right an image generated applying our semantically aware GAN on a synthetic image  from the GTA dataset \cite{richter2016playing} (left) to make the latter look more realistic. Lower right corner: zoomed crops to highlight how our semantically aware GAN can transform images across domains preserving the semantic structure of the scene.}
    \label{fig:teaser}
\end{figure}

Many recent works \cite{richter2016playing,richter2017playing,ros2016synthia,Dosovitskiy17} have proposed to deploy synthetic training images generated by state-of-the-art computer graphics techniques to obtain for free, during the rendering process, different kinds of annotations. Yet, such synthetic training samples turn out significantly different from the real images processed at test time,  which implies a well-known issue, referred to in the machine learning literature as \textit{domain shift}. As a consequence, models trained only on synthetic data severely under-perform in the real deployment scenario. Therefore, the commonly used training protocol mandates the use of (potentially few) annotated training samples from the target domain to perform fine-tuning and recover good performance.
The assumption of having annotated images from the target domain at hand, unfortunately, can not always be fulfilled for complex tasks like 3D reconstruction, semantic segmentation or pose estimation where data acquisition and, especially, labeling is often a challenging and costly task per se. Promising works like \cite{tzeng2017adversarial,ganin2016domain,long15Learning} try to learn models which extract the same kind of features across the two domain. While this strategy seems successful for tasks like classification, it is still unclear how well it can scale to dense \textit{structured domain adaptation} \cite{yamada2014domain} where the improvement gained by feature alignment is still modest. 
Alternatively, \cite{shrivastava2017learning,bousmalis2017using} work directly on the training data trying to shrink the gap between synthetic and real images by transforming the first to make them look real using image-to-image generative adversarial networks. However, since they do not enforce any kind of constraint on the geometric consistency between input and output, these approaches can easily produce artifacts and distortions. Beside harming the realism of the generated images, artifacts could easily render annotations created for the synthetic images useless, especially for pixel-level labeling task where even a few pixels shift can invalidate the annotation.

Building on these observations we propose a novel approach based on image-to-image domain translation by GANs while explicitly training the system to keep the semantic structure of the scene. The intuition behind our formulation is that forcing the \textit{generator} network to keep the semantic structure of the image act as a regularizer enforcing overall consistency of image appearance and producing images that look more realistic and exhibit less artifacts. For example, according to our formulation a "tree" can change its appearance but it should still be recognizable as a "tree" across domains. To enforce the semantic constraint we train a \textit{discriminator} network not only to classify the domain (real/fake) but also to solve the task of semantic segmentation on the synthetic domain (\ie, we do not need labels on the real target domain). Moreover, we introduce an appearance reconstruction loss to further regularize the generation process and help preserving small details across the domain adaptation. 
To asses the effectiveness of our proposal we transform synthetic images obtained from the synthetic GTA datasets \cite{richter2016playing} to look similar to the real images of the Cityscapes \cite{Cordts_2016_CVPR} dataset. \autoref{fig:teaser} shows on the right column a qualitative example generated by our method using as input the corresponding synthetic images depicted on the left. We will show how those images can be used to train a model to solve the problem of semantic segmentation yielding promising result with respect to using synthetic images.

\begin{figure*}[t]
\centering\includegraphics[width=0.80\textwidth]{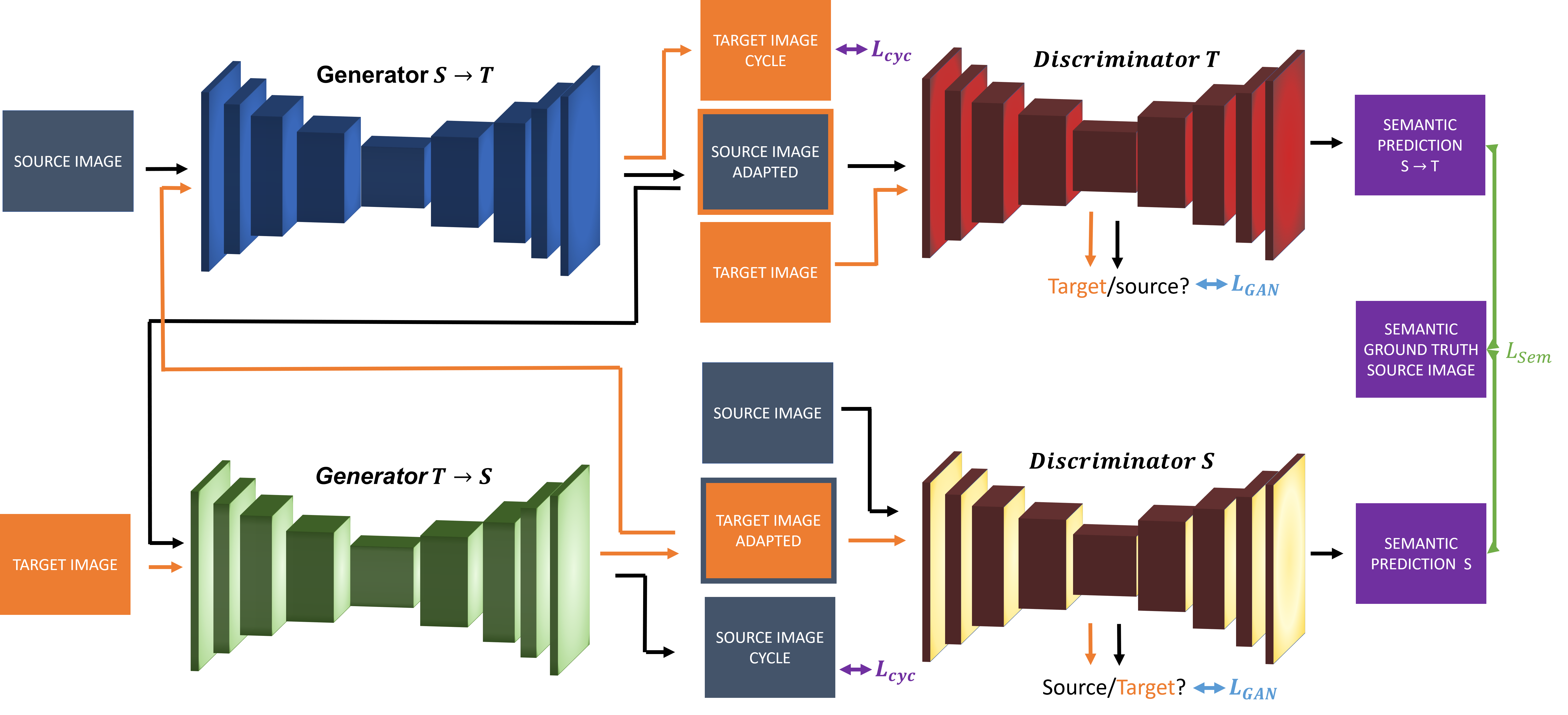}
\caption{Schematic representations of the proposed network architecture. In dark blue and orange images from the source domain and target domain respectively. Dual color framed images are obtained by our adaptation method. In purple the semantic maps. 
}
\label{fig:netowork}
\end{figure*}

\section{Related works}
\label{sec:related}
The main topics relevant to this article are Semantic Segmentation and Domain Adaptation.

\subsection{Semantic Segmentation}
Since the advent of deep learning, semantic segmentation is mainly performed by convolutional neural networks\cite{long15Learning}. Several kind of architectures are employed. Multi-scale models, such as \cite{eigen2015predicting,chen2016attention,chen2018deeplab,liang2015semantic}, take inputs at different resolution to extract context information at different abstraction levels.  Encoder-decoder networks, \cite{long2015fully,badrinarayanan2017segnet,ronneberger2015u}, combines an encoder to extract high-level, low-resolution features which are later exploited by a decoder to reconstruct an high-resolution semantic map. Other networks employ Conditional Random Fields,  \cite{chen2018deeplab,zheng2015conditional}, to encode long range context information. Modern networks deploy spatial pyramid pooling and atrous-convolutions to extract information at different level of abstraction \cite{zhao2017pyramid,chen2018deeplab,deeplabv3plus2018}.

\subsection{Domain Adaptation}
Many domain adaptation techniques have been proposed to address the domain-shift problem \cite{patel2015visual,wang2018deep,csurka2017domain}. Earliest approaches such as \cite{gong2012geodesic,gopalan2011domain} try to build intermediate representations using manifolds while recent ones, tailored for deep learning, focus on adversarial training. Deep domain adaptation can be mainly divided in two branches: Pixel-level and Feature-Level. Feature-level approaches, such as \cite{long15Learning,liu2016coupled,tzeng2017adversarial,ganin2016domain} seek to find a domain invariant representation, obtaining networks able to perform well on both the source and target domains. On the other hand, pixel-level approaches, such as \cite{shrivastava2017learning}, work on image data and try to directly convert the source image into a target style image relying on recent  image-to-image translation generative networks \cite{Isola_2017_CVPR,Zhu_2017_ICCV}. 
Few works have explicitly studied domain adaptation for semantic segmentation. \cite{hoffman2016fcns} performs two kind of alignments: a global alignment through adversarial back-propagation, as in \cite{ganin2015unsupervised}, and a local one, which aligns class specific statistics by a multiple instance learning formulation. \cite{zhang2017curriculum} proposes curriculum-style learning where a teacher network solve the easier task of learning global label distributions over images and local distributions over landmark superpixels, then a student segmentation network is trained so that the target label distribution follow these inferred label properties. 
Similarly to our proposal, \cite{Hoffman_cycada2017} combines the cycle consistency loss proposed by \cite{Zhu_2017_ICCV} with a semantic consistency loss.
While they combine different networks trained sequentially, 
in our work we proposed a simpler end-to-end architecture with a semantic discriminator that obtains comparable or even better results. 

\section{Proposed Method}
\label{sec:method}
In this section we present our proposal for domain adaptation exploiting semantic information. We consider the problem of unsupervised and unpaired pixel-level domain adaptation from a source to a target domain. We define as $X_s$ , $Y_s$ the provided source data and associated semantic labels whilst as $X_t$ the provided target data, but without any available target labels. Our goal is to transform source images so to resemble target images while maintaining the semantic content of the scene during the generation process. A schematic representation of our method is shown in \autoref{fig:netowork}.

\subsection{Architecture}
Inspired by  \cite{Zhu_2017_ICCV}, we adopt a cycle architecture consisting of two generators and two semantic discriminators. The first generator, $G_{S \rightarrow T}$, introduce a mapping from source to target domain and produces target samples which should deceive the discriminator $D_T$. The discriminator $D_T$, instead, learns to distinguish between adapted source and true target samples. On the other hand, the second generator, $G_{T \rightarrow S}$, learns the opposite mapping from source to target data, while the second discriminator $D_s$ distinguish between adapted target and true source samples. 
Furthermore, peculiarly to our work, both \textit{semantic} discriminators act not only as classifiers but also as semantic segmentation networks. Thus, we add a second decoder to $D_T$ and $D_S$ obtaining $D_{S_{sem}}$ and $D_{T_{sem}}$. The features extracted by the last encoder layer of the discriminators are used to generate both the semantic map and the domain classification score.

\subsection{Training}
\label{sec:losses}
We train our system to minimize multiple losses: \\

\textbf{Adversarial Loss}\\
We apply adversarial losses \cite{goodfellow2014generative} to both mapping functions $S \rightarrow T$ and $T \rightarrow S$. For the sake of space, we define here only source to target adversarial loss, being equivalent to its inverse.

\begin{equation}\begin{split}
\mathcal{L}_{adv} = \EX_{\mathrm{x_t \sim X_T}}[log(D_t(x_t))] + \\
\EX_{\mathrm{x_s \sim X_s}}[log(1 - D_t(G_{s \rightarrow t}(x_s)))]
\end{split}
\label{eq:loss_adversarial}
\end{equation}

$G_{S \rightarrow T}$ tries to generate images that look similar to images from domain $T$ while $D_T$ tries to distinguish between adapted source samples $G_{S \rightarrow T}(x_S)$ and real target samples $X_T$. $G_{S \rightarrow T}$ seek to minimize this objective against $D_T$ which instead tries to maximize it.

\textbf{Semantic Discriminator Loss}\\
We train both discriminators, $D_{S_{sem}}$ and $D_{T_{sem}}$, to perform semantic segmentation employing source labels. $D_{T_{sem}}$ will be trained on adapted source images, while $D_{S_{sem}}$ will be trained directly on source images. We used a pixel-wise cross entropy loss $H(p,q)$ as in standard segmentation networks:

\begin{equation}\begin{split}
\mathcal{L}_{sem} = H(D_{S_{sem}}(G_{S \rightarrow T}(X_S)), Y_S) + H(D_{S_{sem}}(X_S), Y_S)
\end{split}
\label{loss_sem}
\end{equation}

\textbf{Weighted Reconstruction Loss}\\
We exploit the cyclic L1 reconstruction loss proposed in \cite{Zhu_2017_ICCV} for target samples where we do not have any label. 
Regarding source samples, we weight each pixel proportionally to the probability of not belonging to its semantic class. Our weighting term acts as a regularization where the network usually fail adaptation introducing artifacts, forcing the least frequent classes to be reconstructed preserving input appearance:

\begin{equation}\begin{split}
    \mathcal{L}_{rec} = ||G_{S \rightarrow T}(G_{T \rightarrow S}(x_T))) - x_T||_1 + \\
    (1-w)||G_{T \rightarrow S}(G_{S \rightarrow T}(x_S))) - x_S||_1
\end{split}\end{equation}

$w$ is a weight mask with the same resolution of the source image. Defined $C$ as the set of possible classes, each weight $w_{i,j}$ represents the likelihood of a class among all synthetic dataset: 

\begin{equation}
    w_{i,j} = \frac{n_{pixel} \in c}{n_{pixel}}, c \in C
\end{equation}.

\textbf{Final Loss}\\
We train our discriminators and generators to minimize the following losses:

\begin{equation}\begin{split}
    \mathcal{L}_{D} = -\mathcal{L}_{adv} + \lambda_{sem}\mathcal{L}_{sem}\\
    \mathcal{L}_{G} = \mathcal{L}_{adv} + \lambda_{sem}\mathcal{L}_{sem} + \lambda_{rec}\mathcal{L}_{rec}\\
\end{split}\end{equation}

$\lambda_{sem}$ and $\lambda_{rec}$ are hyper-parameters that control the relative importance of domain classification, weighted reconstruction and semantic segmentation. Across all our experiments we will use $\lambda_{sem}$  = 1 and $\lambda_{rec}$= 3. 

\section{Experimental Results}
\label{sec:experimental}

\begin{figure*}
\centering
\small
\setlength{\tabcolsep}{1pt}
\begin{tabular}{cccc}
(a) GTA & (b) GTA $\rightarrow$ Cityscapes \cite{Zhu_2017_ICCV} & (c) GTA $\rightarrow$ Cityscapes (\textit{ours}) & (d) Cityscapes \\ 
\includegraphics[width=0.21\textwidth]{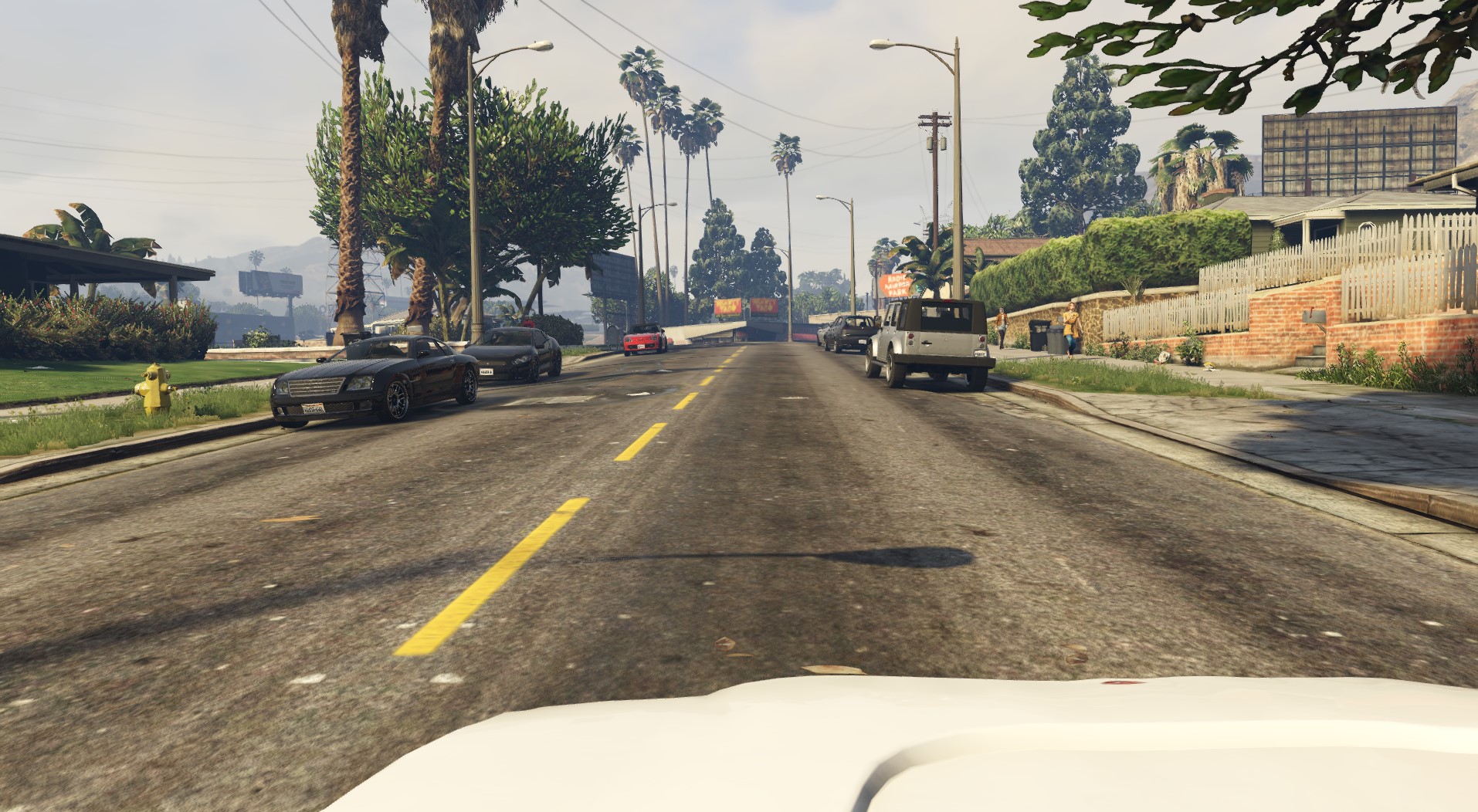} & \includegraphics[width=0.21\textwidth]{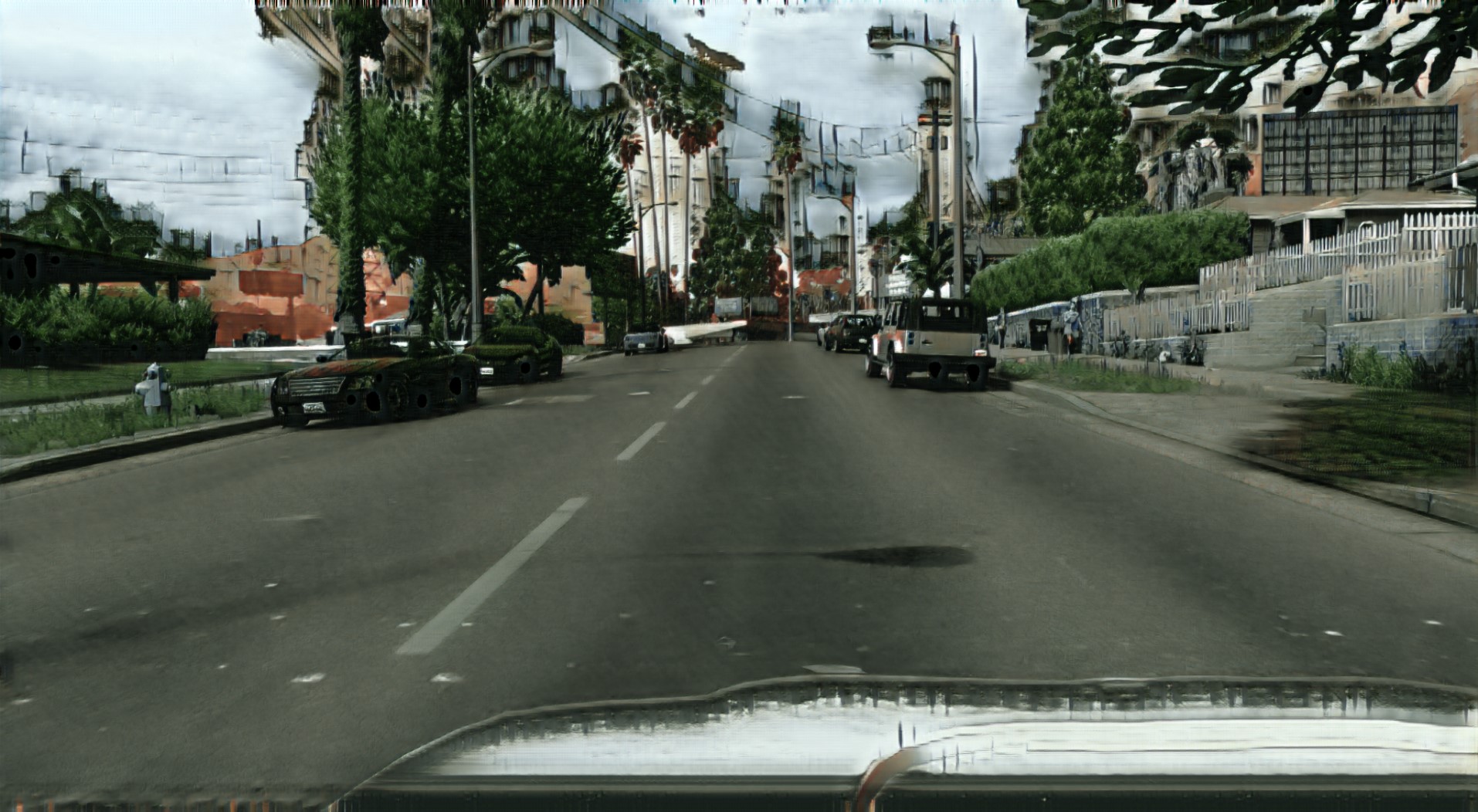} &
\includegraphics[width=0.21\textwidth]{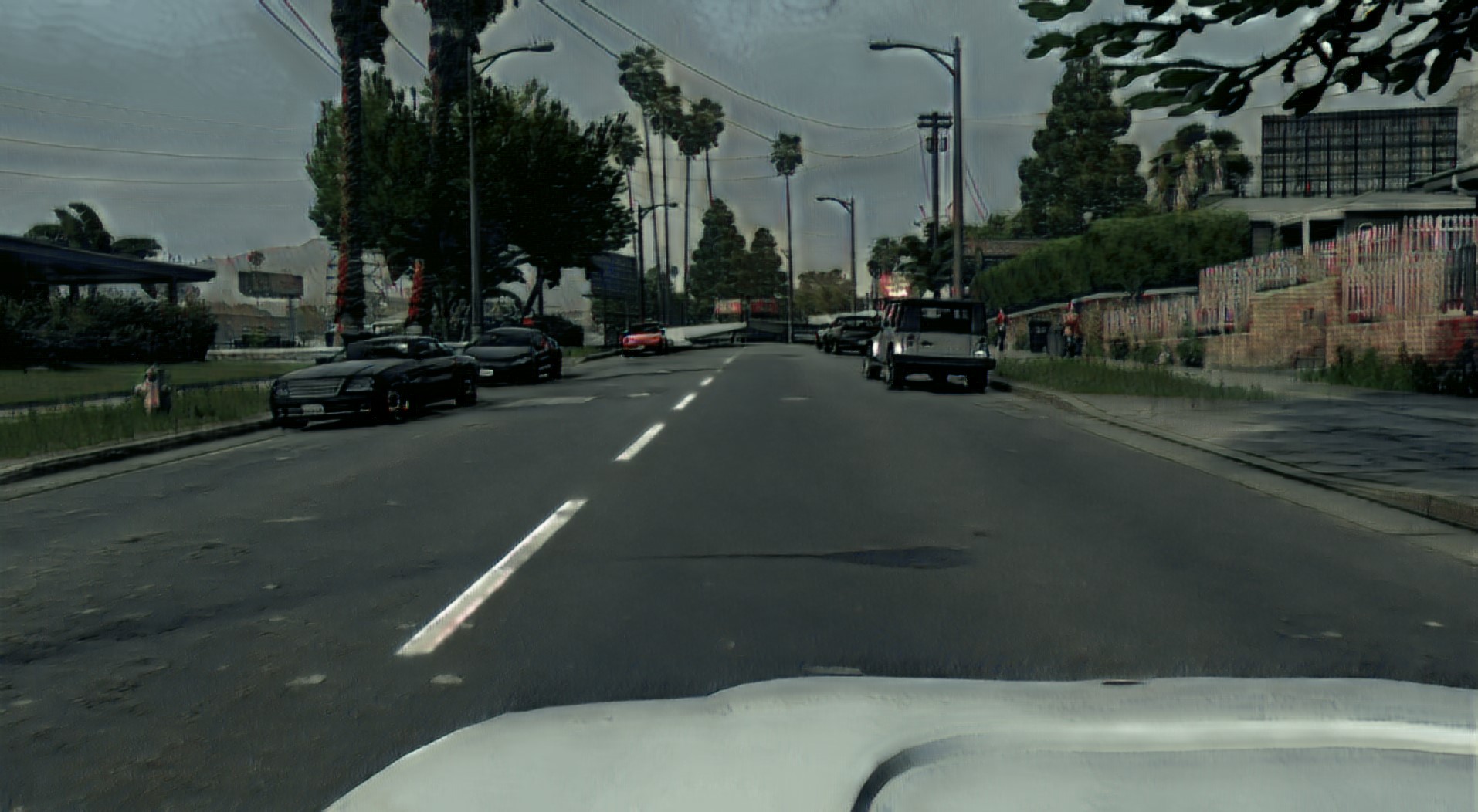} &
\includegraphics[width=0.21\textwidth]{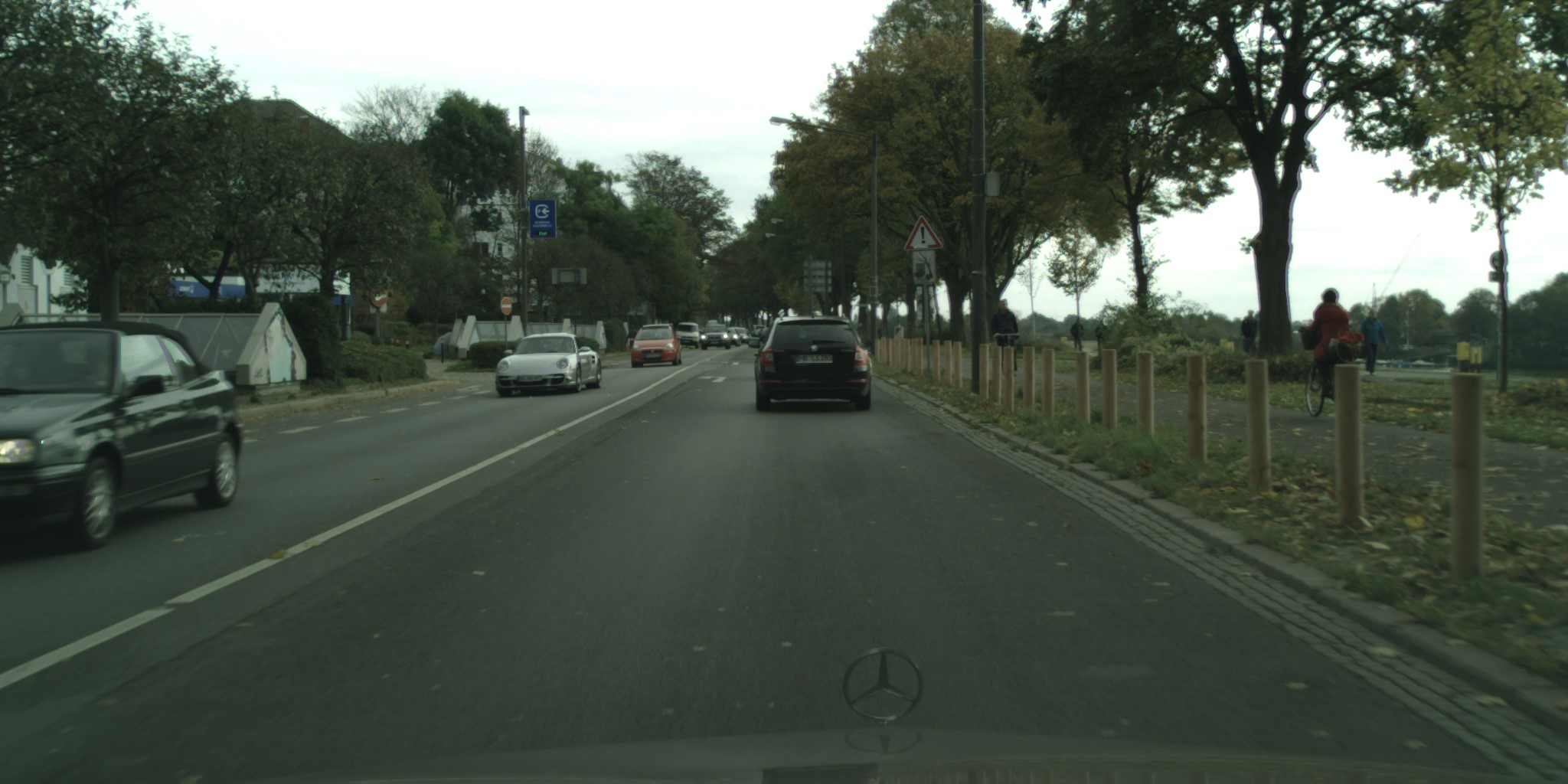} \\

\includegraphics[width=0.21\textwidth]{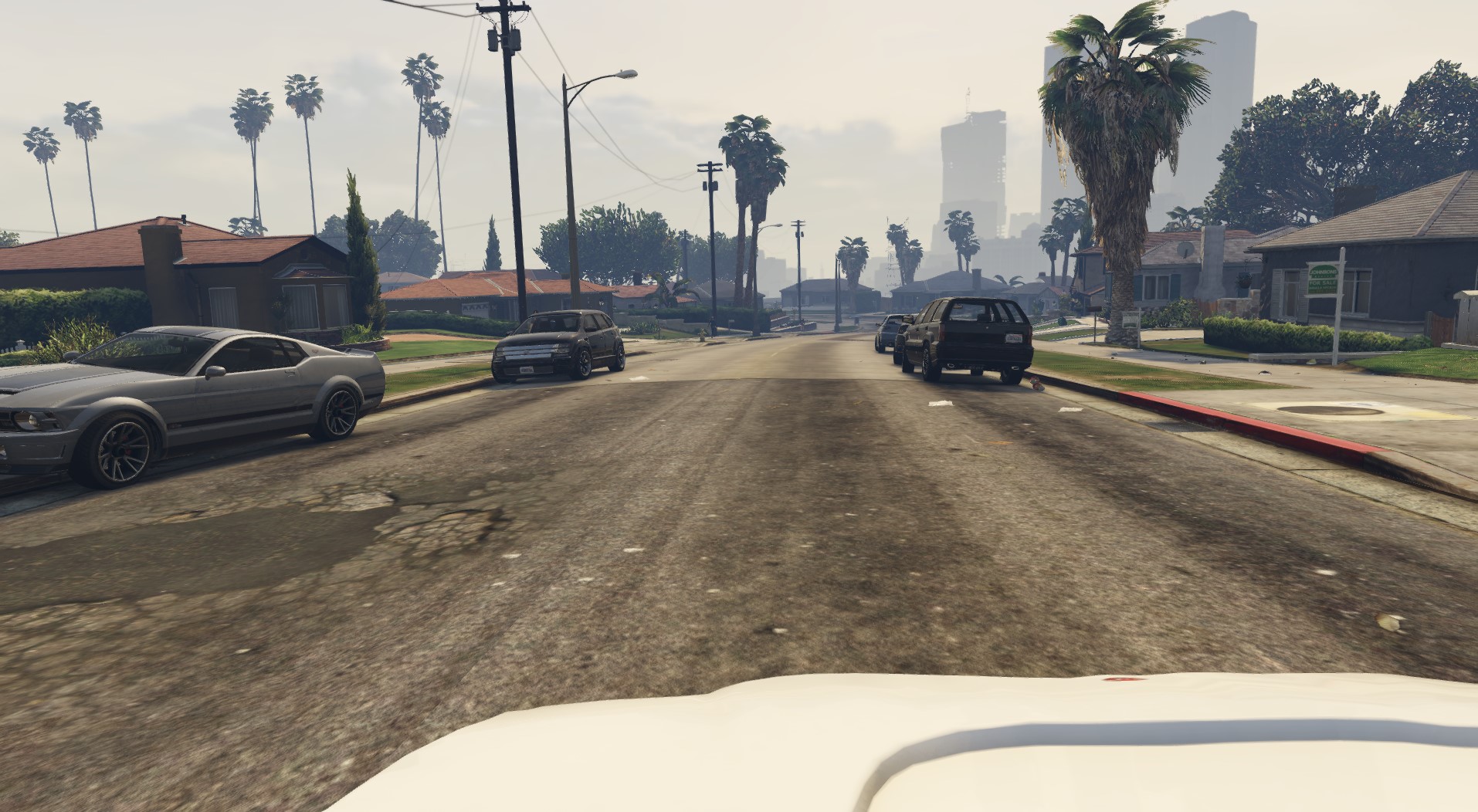} & \includegraphics[width=0.21\textwidth]{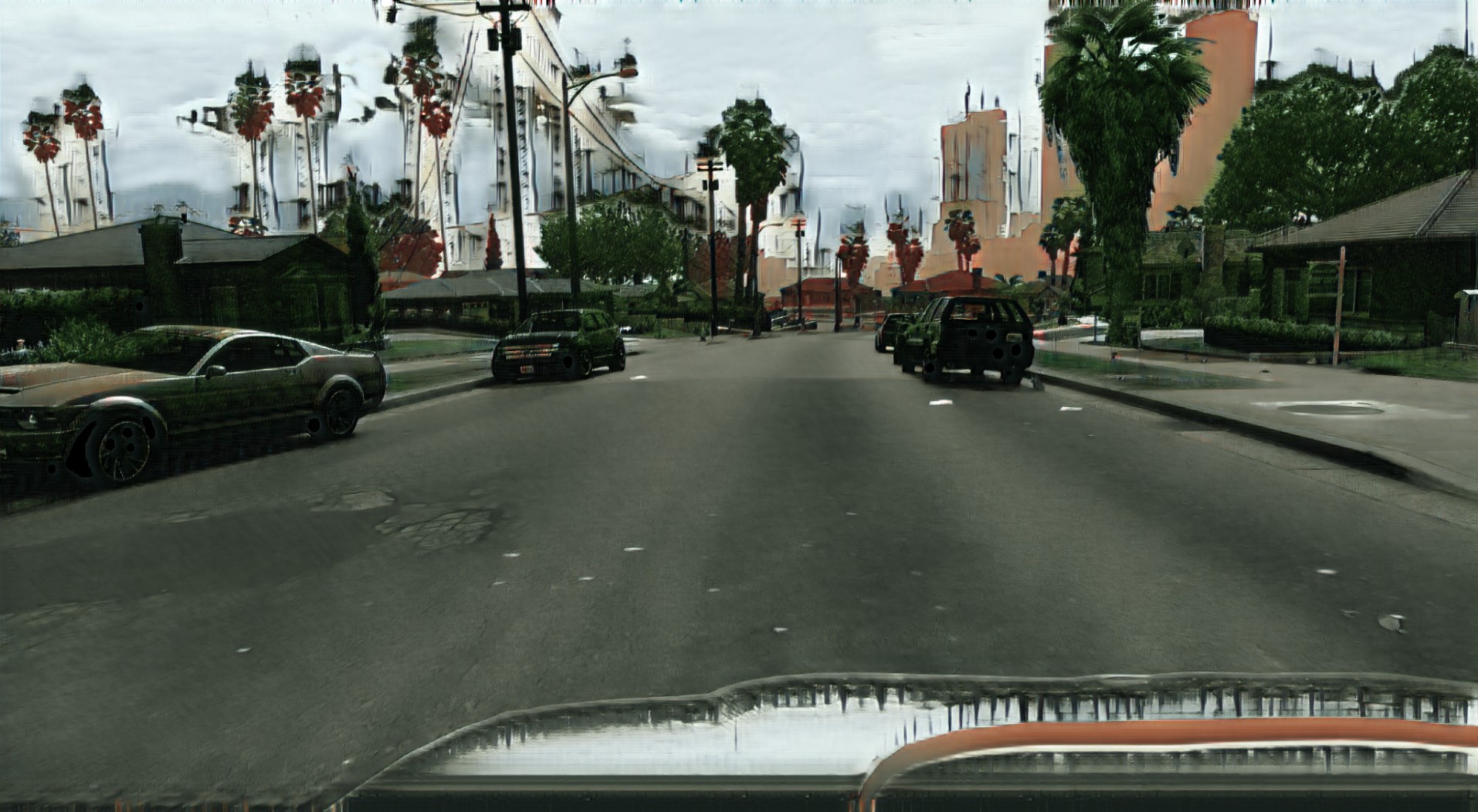} &
\includegraphics[width=0.21\textwidth]{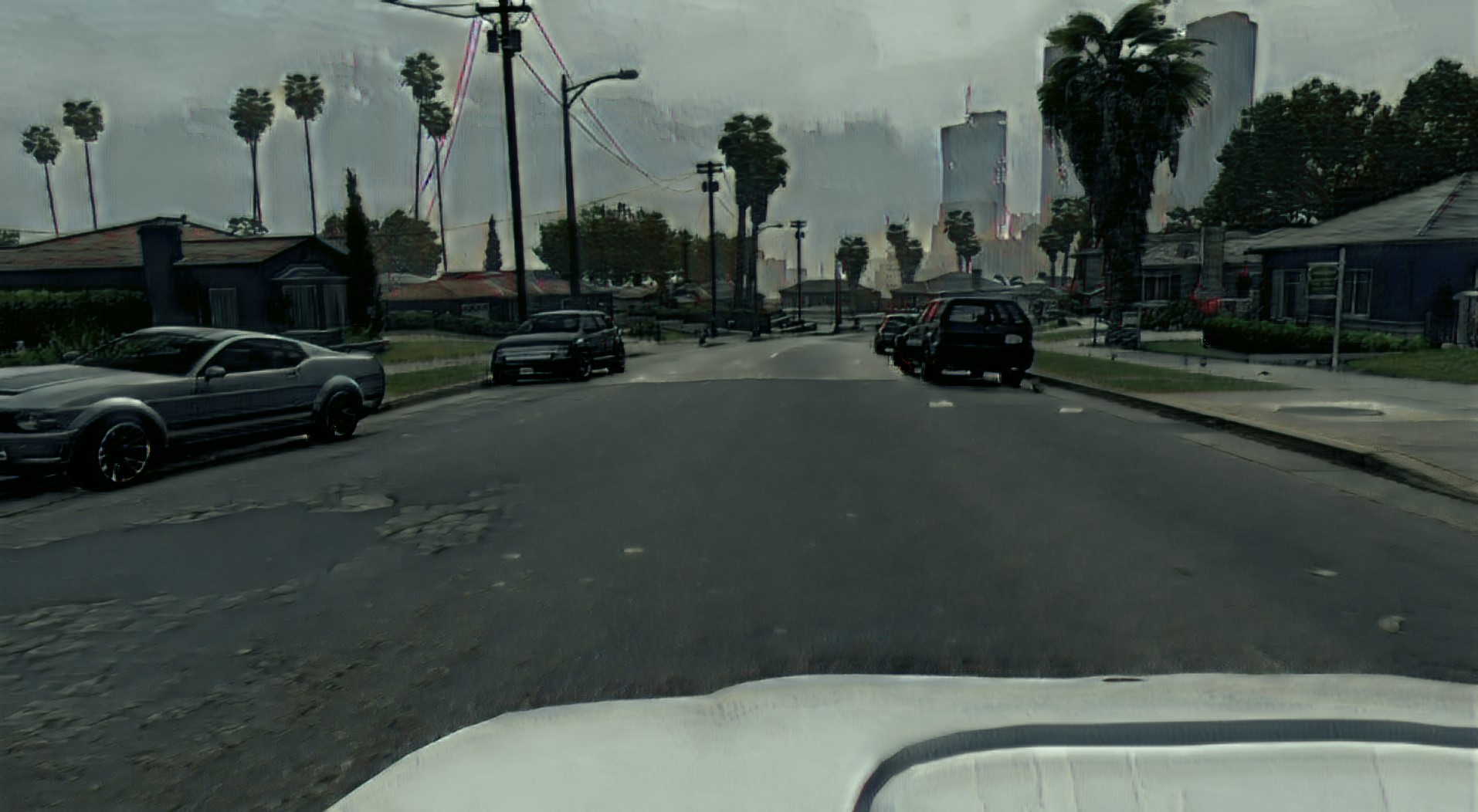} &
\includegraphics[width=0.21\textwidth]{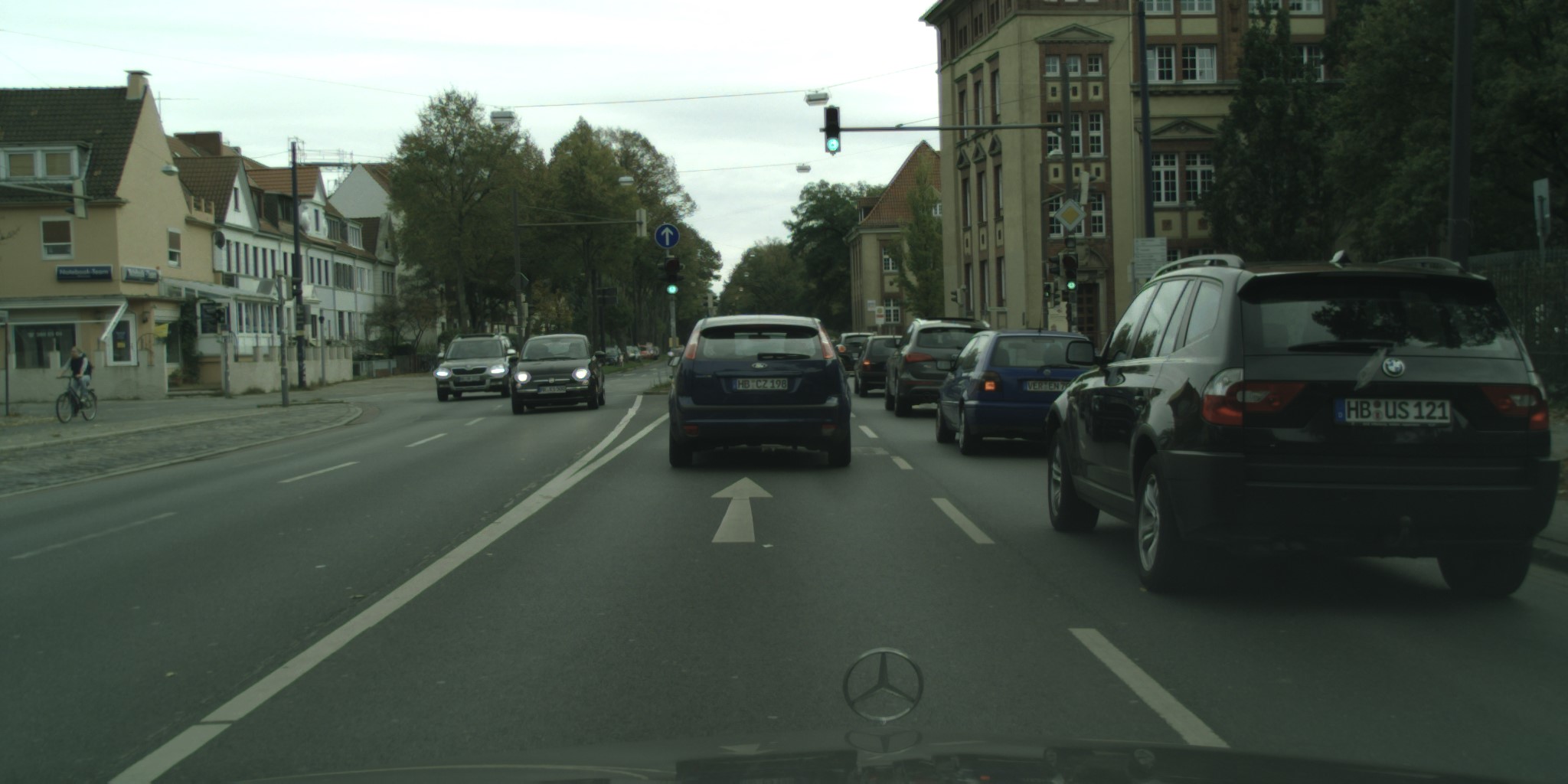} \\

\end{tabular}
\caption{Image generated by CycleGan \cite{Zhu_2017_ICCV}(b) and  our semantics-aware GAN (c) for the  GTA to Cityscapes domain alignement task}
\label{fig:adapted}
\end{figure*}

We conduct a series of tests to assess the effectiveness of our method in producing realistic images and verify if they are suitable for training deep learning models. 
\subsection{Datasets Creation}
We have used as synthetic source domain the GTA dataset \cite{richter2016playing}, that features 22K realistic synthetic images obtained from the Grand Theft Auto videogame enriched with perfect pixel level annotations for semantic segmentation. As target real images we have used the Cityscapes dataset \cite{Cordts_2016_CVPR} featuring 5000 images acquired during real driving sessions around Germany and annotated with precise pixel level labels for semantic segmentation. Among all available images we have used the \textit{training} split as our target samples during training, while we have kept the \textit{validation} split to measure performance of different semantic segmentation networks. We did not use the \textit{test} split since the labels are not publicly available.  
We chose these two datasets as they provide annotations for the same set of semantic classes and feature domains where the biggest difference concern the shift from synthetic to real images.
We used ResNet as our generator networks and U-Net \cite{ronneberger2015u} as our discriminator. 
Using the loss formulation described in \autoref{sec:losses} we have trained our GAN to transform images from the GTA \cite{richter2016playing} to the Cityscapes \cite{Cordts_2016_CVPR} domain for 300k steps using Adam as optimizer, 0.0001 for learning rate and batch size 2. We cropped our input images to 512x512. During the training process we have used images and labels from GTA and only images from Cityscapes, \ie, our method does not require annotations from the real/target domain but only from the source one. Once trained, we used the generator to transform synthetic images from the training dataset to produce an \textit{aligned} GTA dataset that should resemble images from the real Cityscapes domain. On \autoref{fig:adapted} we depict some qualitative example of images produced by our GAN (column (c)) together with the corresponding input from the GTA dataset (column (a)) and some exemplar images from the Cityscapes, target, dataset (column (d)). To better show the effectiveness of our semantic aware GAN, we also report images obtained by training a CycleGAN network \cite{Zhu_2017_ICCV} that does not use any semantic clues at training time  (column (b)). By comparing our images (column (c)) with those produced by CycleGAN (column (b)) it turns out clearly that, unlike previous approaches (\ie, column (b)), our novel formulation can preserve the semantic content and avoid introduction of artifacts. Moreover, the introduction of semantic constraints during the training process helps to produce sharper edges in the final image, which increases the quality  of the images compared to CylceGAN. We have also applied our GANs to entire video sequences from the GTA domain and verified that the network can easily maintain temporal consistency even if it has only been trained on single frames. \footnote{\url{https://youtu.be/wIpFcKLviYQ}}

\begin{table*}[t]
\center
\setlength{\tabcolsep}{2.5pt}
\begin{tabular}{|c|ccccccccccccccccccc|cc|}
    \hline
    Method & \rotatebox{90}{road} & \rotatebox{90}{sidewalk} & \rotatebox{90}{building} & \rotatebox{90}{wall} & \rotatebox{90}{fence} & \rotatebox{90}{pole} & \rotatebox{90}{traffic light} & \rotatebox{90}{traffic sign} & \rotatebox{90}{vegetation} & \rotatebox{90}{terrain} & \rotatebox{90}{sky} & \rotatebox{90}{person} & \rotatebox{90}{rider} & \rotatebox{90}{car} & \rotatebox{90}{truck} & \rotatebox{90}{bus} & \rotatebox{90}{train} & \rotatebox{90}{motorcycle} & \rotatebox{90}{bicycle} & \rotatebox{90}{mIoU} & \rotatebox{90}{Acc.} \\
    \hline 
    Source \cite{hoffman2016fcns} & 31.9 & 18.9 & 47.7 & 7.40 & 3.10 & 16.0 & 10.4 & 1.00 & 76.5 & 13.0 & 58.9 & 36.0 & 1.00 & 67.1 & 9.50 & 3.70 & 0.00 & 0.00 & 0.00 & 21.2 & -\\
    \cite{hoffman2016fcns} & 70.4 & 32.4 & 62.1 & 14.9 & 5.40 & 10.9 & 14.2 & 2.70 & 79.2 & 21.3 & 64.6 & 44.1 & 4.20 & 70.4 & 8.00 & 7.30 & 0.00 & 3.50 & 0.00 & 27.1 & - \\
    \hline
    Source \cite{zhang2017curriculum} & 18.1 & 6.80 & 64.1 & 7.30 & 8.70 & 21.0 & 14.9 & 16.8 & 45.9 & 2.40 & 64.4 & 41.6 & 17.5 & 55.3 & 8.40 & 5.0 & 6.90 & 4.30 & 13.8 & 22.3 & -\\ 
    \cite{zhang2017curriculum} & 74.9 & 22.0 & 71.7 & 6.00 & 11.9 & 8.40 & 16.3 & 11.1 & 75.7 & 13.3 & 66.5 & 38.0 & 9.30 & 55.2 & \textbf{18.8} & 18.9 & 0.00 & \textbf{16.8} & \textbf{16.6} & 28.9  & -\\
    \hline
    Source \cite{Hoffman_cycada2017} & 26.0 & 14.9 & 65.1 & 5.50 & 12.9 & 8.90 & 6.00 & 2.50 & 70.0 & 2.90 & 47.0 & 24.5 & 0.0 & 40.0 & 12.1 & 1.50 & 0.0 & 0.0 & 0.0 & 17.9 & 54.0 \\
    \cite{Hoffman_cycada2017} & 83.5 & \textbf{38.3} & 76.4 & 20.6 & \textbf{16.5} & 22.2 & \textbf{26.2} & \textbf{21.9} & 80.4 & \textbf{28.7} & 65.7 & 49.4 & 4.2 & 74.6 & 16.0 & \textbf{26.6} & 2.0 & 8.0 & 0.0 & \textbf{34.8} & 82.8 \\
    \hline
    Source Ours & 43.3 & 11.9 & 54.3 & 3.42 & 11.96 & 9.63 & 10.74 & 5.23 & 68.3 & 6.39 & 46.84 & 30.02 & 2.07 & 33.1 & 7.72 & 0.00 & 0.00 & 0.00 & 0.00 & 18.2 & 60.4 \\
    Ours & \textbf{85}.4 & 32.8 & \textbf{78.0} & \textbf{21.0} & 9.35 & \textbf{26.1} & 18.0 & 8.71 & \textbf{82.2} & 22.1 & \textbf{71.2} & \textbf{51.4} & \textbf{13.4} & \textbf{79.5} & 16.0 & 13.5 & \textbf{7.83} & 10.1 & 0.03 & 34.2 & \textbf{84.4} \\
    \hline
\end{tabular}
\captionsetup{size=small,skip=0.333\baselineskip}
\small
\caption{Comparison between domain adaptation methods for semantic segmentation on the Cityscapes validation set. Middle section reports mIoU score per class, final two columns aggregated performance across the whole dataset, best results highlighted in bold.}
\label{tab:comparison}
\end{table*}

\subsection{Semantic Segmentation}
\label{sec:segmentation_result}

\begin{figure*}[t]
\setlength{\tabcolsep}{1pt}
\centering
\small
\begin{tabular}{cccc}
(a) RGB input & (b) GT segmentation & (c) GTA Trained & (d) GTA $\rightarrow$ Cityscapes Trained\\
\includegraphics[width=0.21\textwidth]{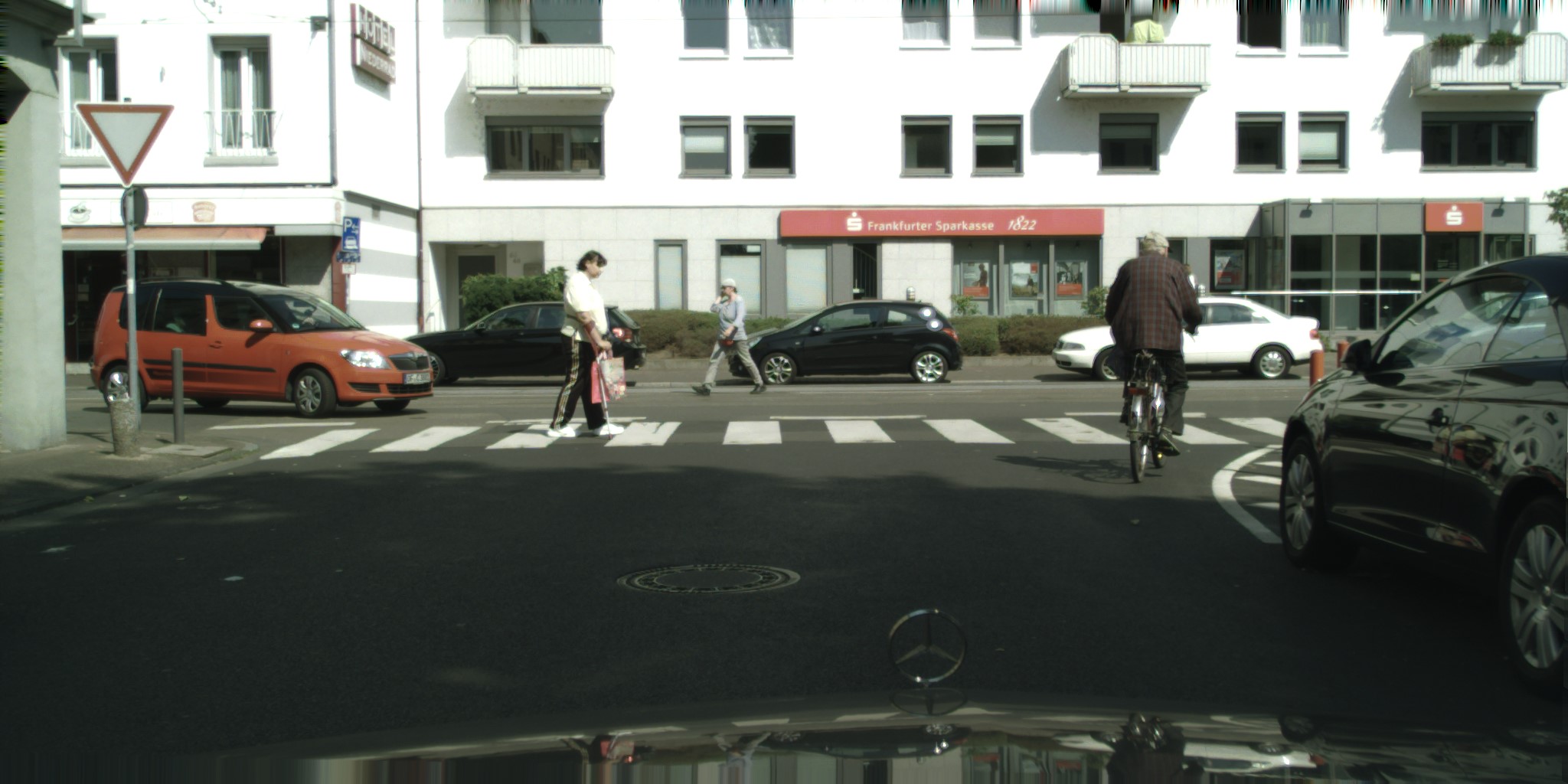} &
\includegraphics[width=0.21\textwidth]{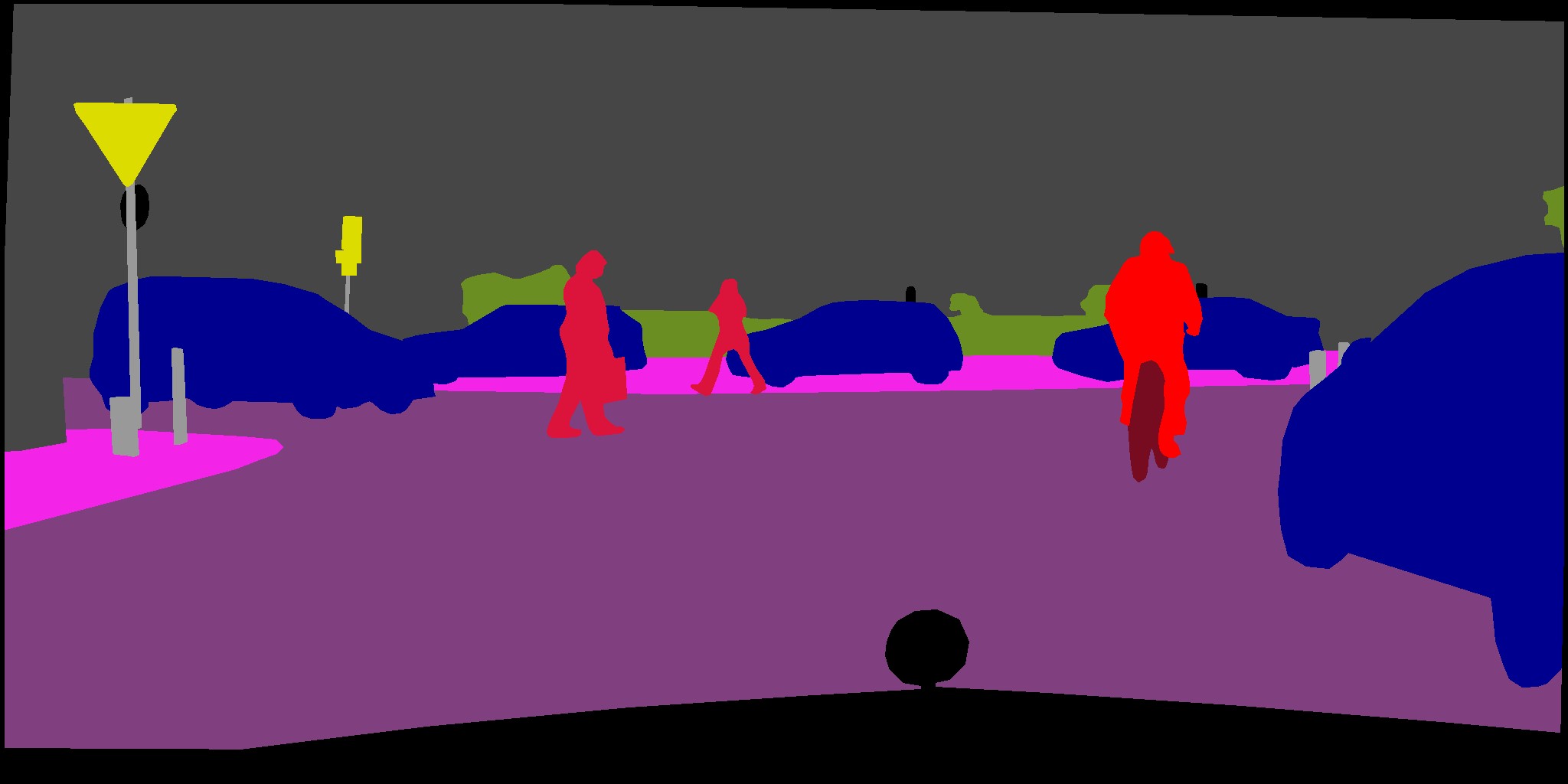}&
\includegraphics[width=0.21\textwidth]{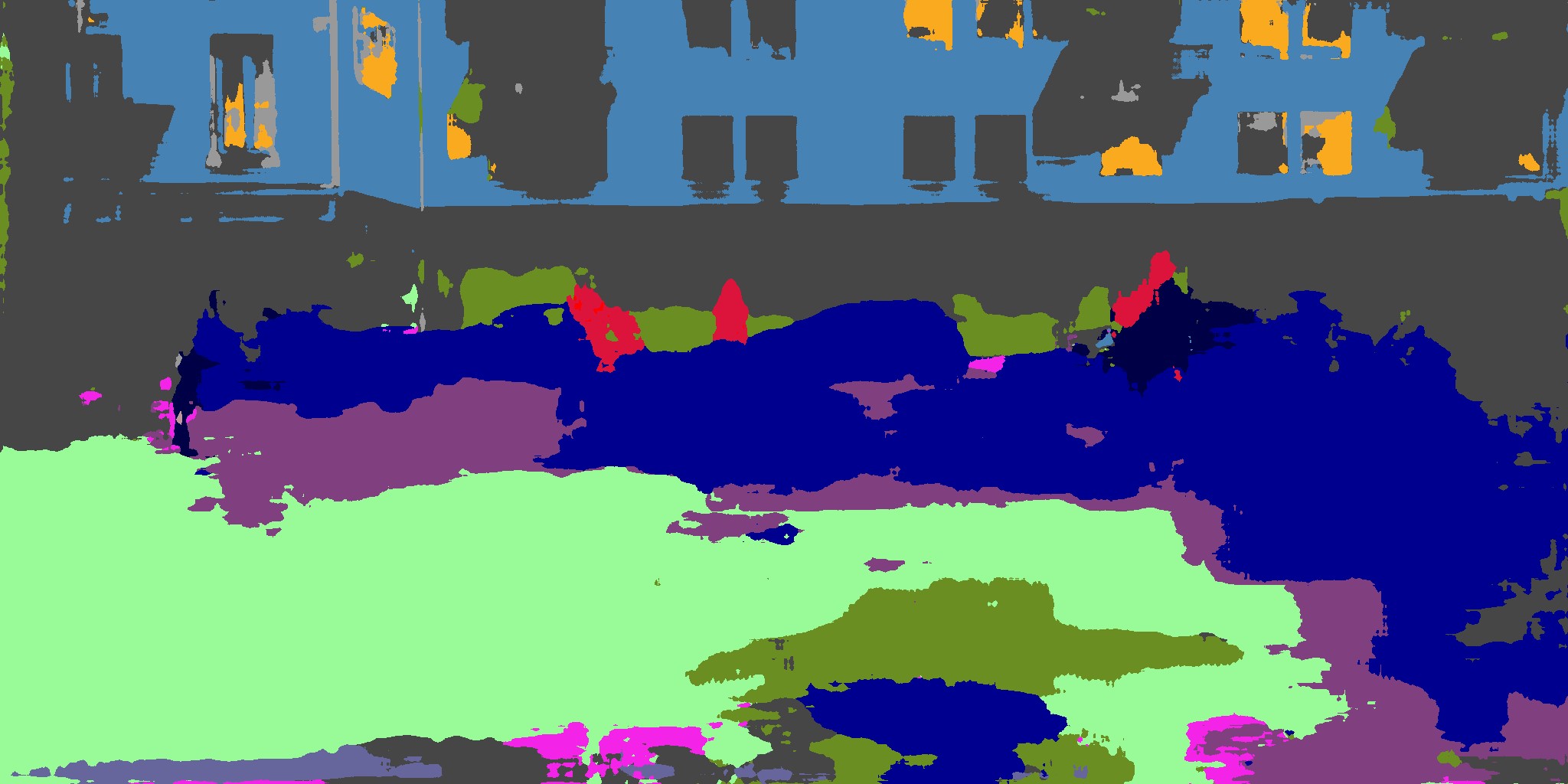} &
\includegraphics[width=0.21\textwidth]{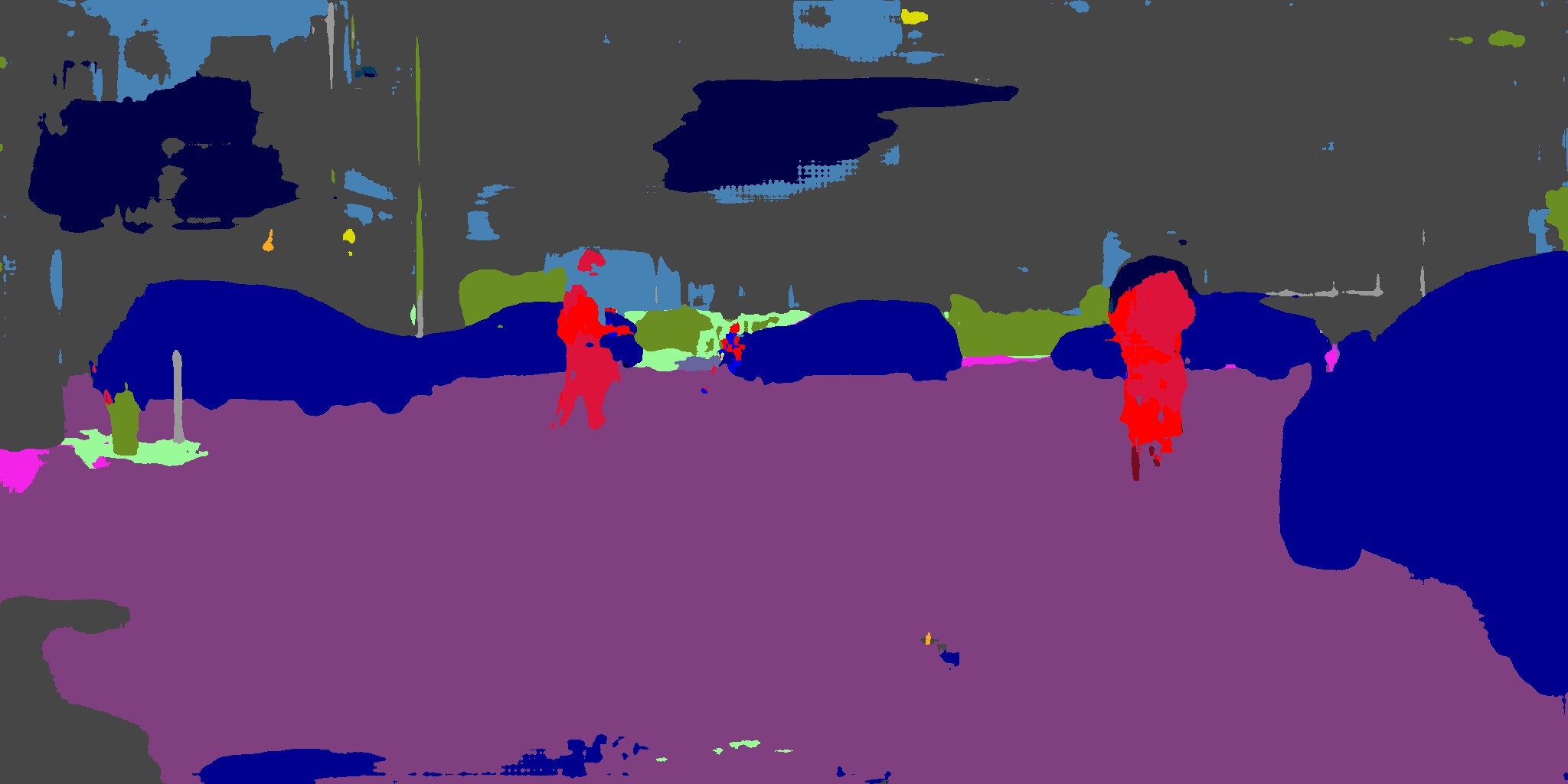} \\

\includegraphics[width=0.21\textwidth]{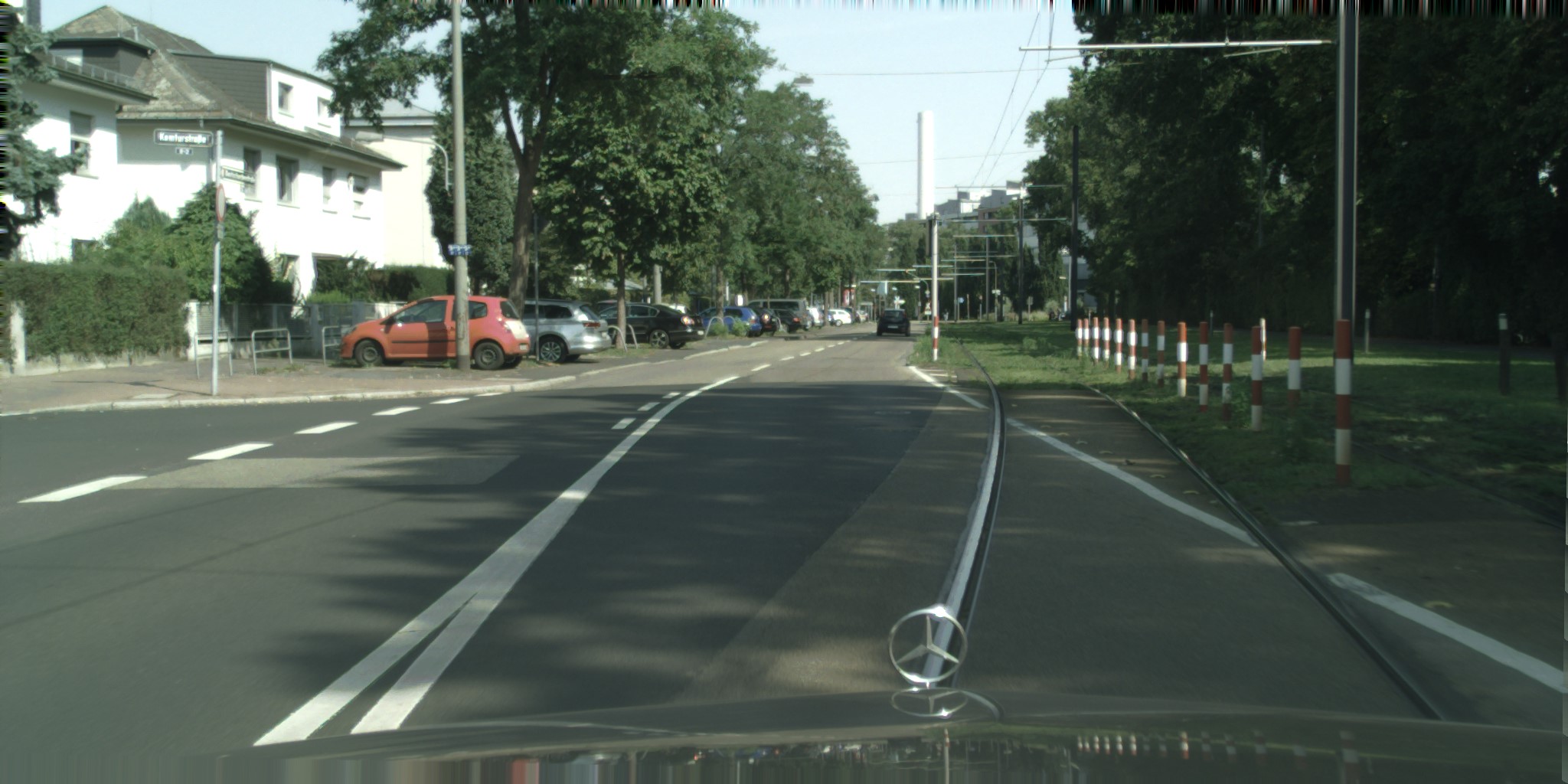} &
\includegraphics[width=0.21\textwidth]{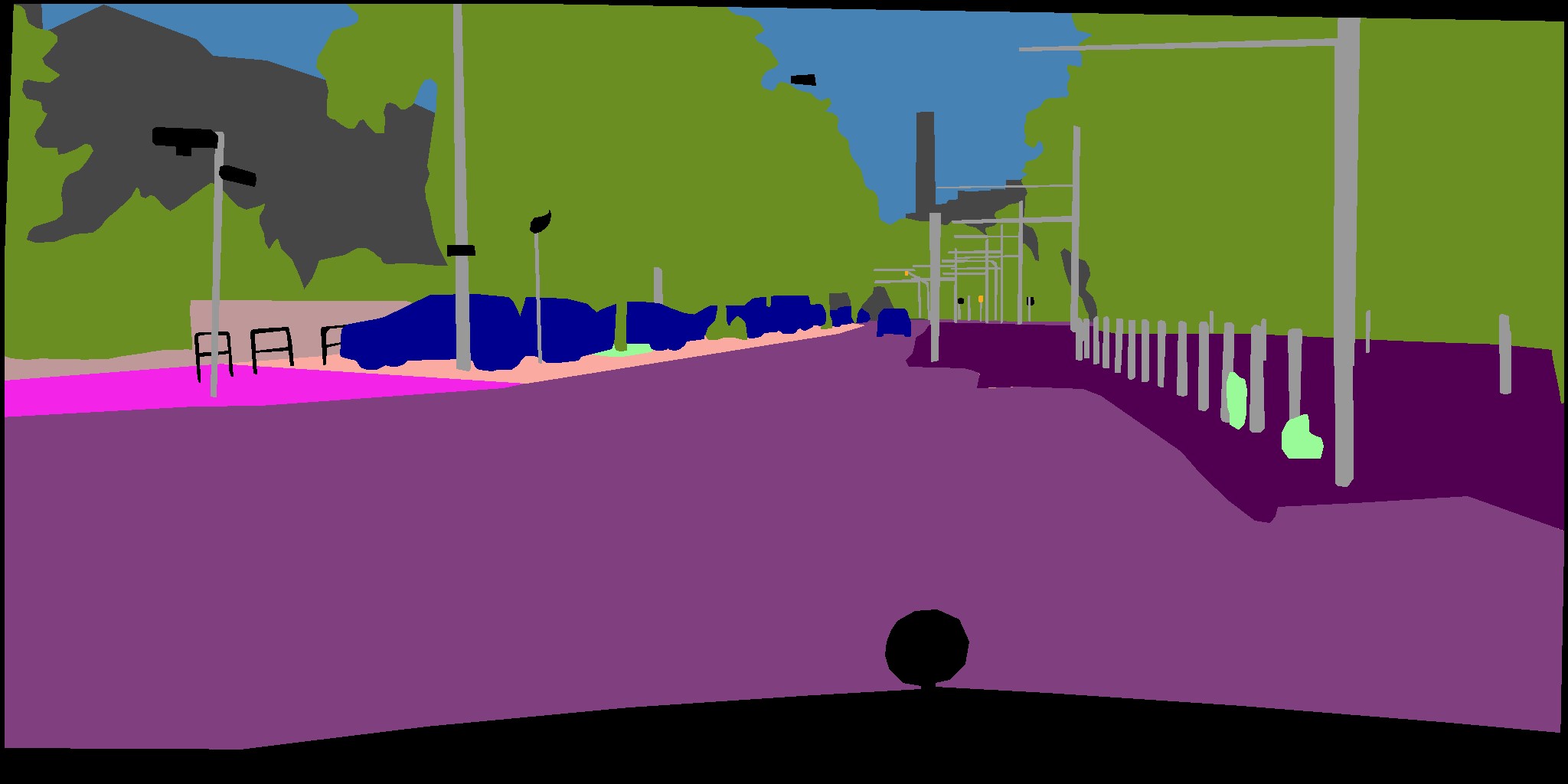} &
\includegraphics[width=0.21\textwidth]{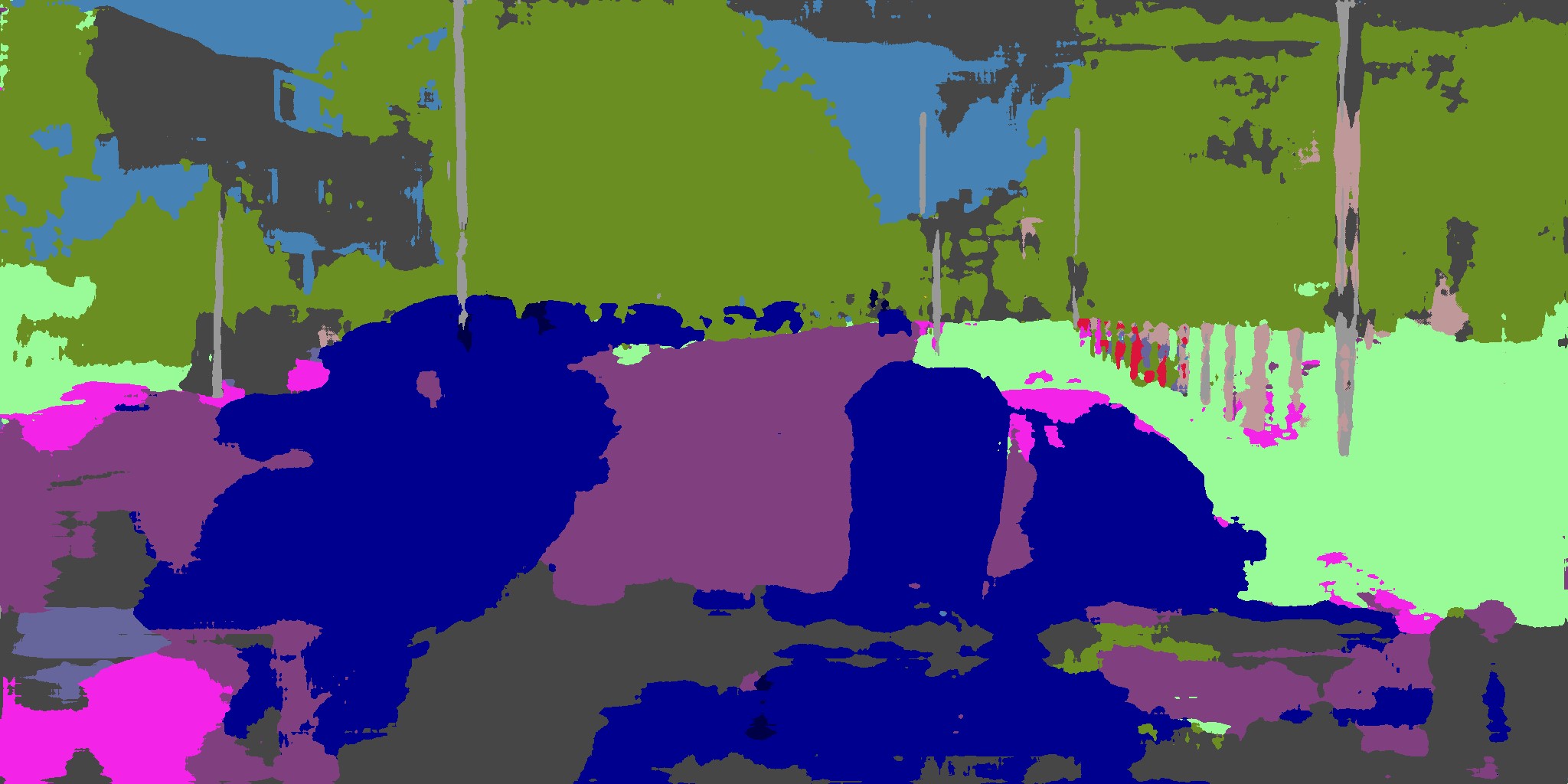} &
\includegraphics[width=0.21\textwidth]{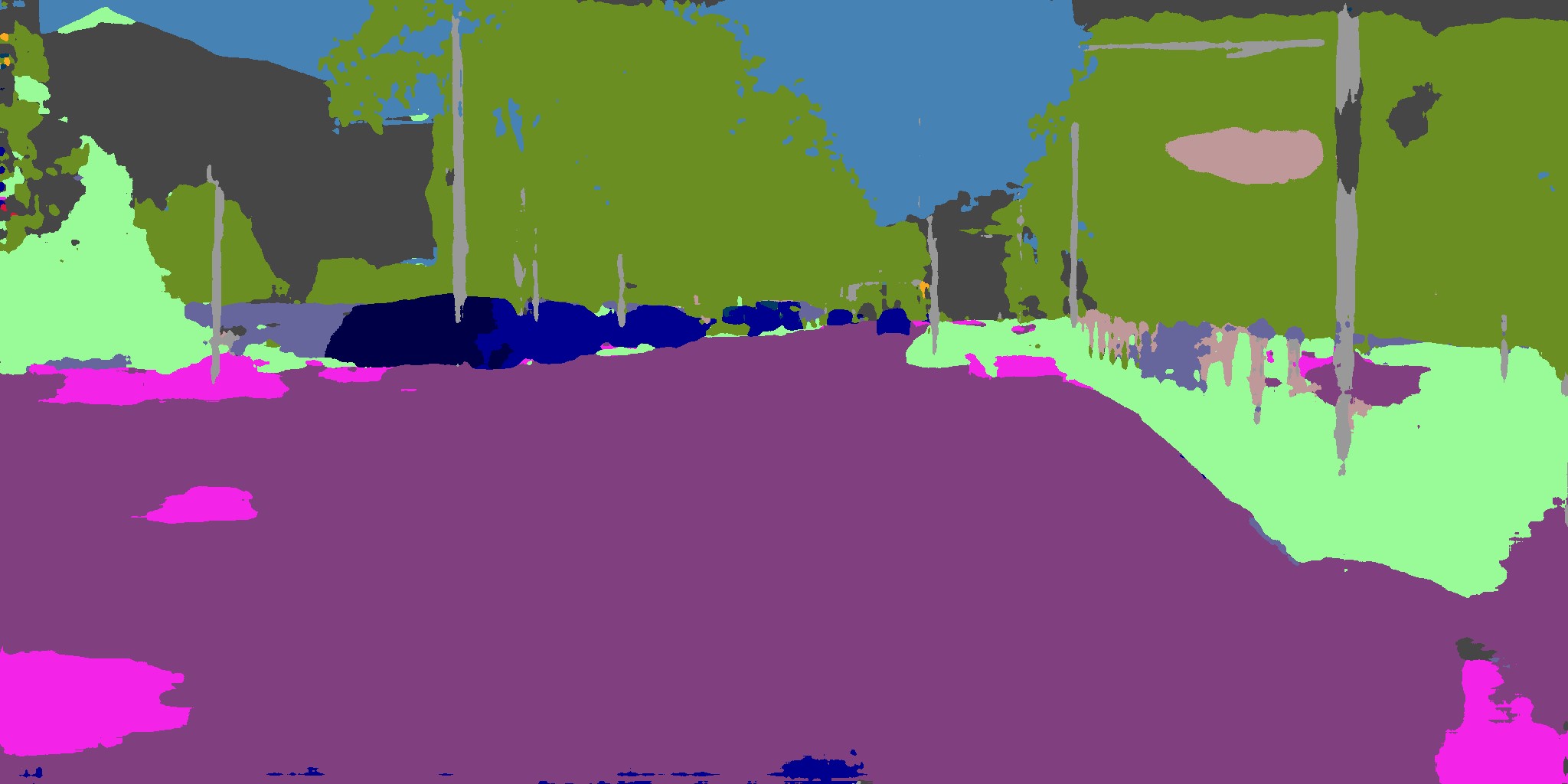} \\
\end{tabular}
\caption{Segmentation results on the Cityscapes dataset for a FCN8s network trained only on synthetic data from the GTA dataset (c) and  on our GTA \textit{adapted} dataset (d).}
\label{fig:qualitative_segmentation}
\end{figure*}

\autoref{fig:adapted} shows how our network can produce visually appealing images. In the following we demonstrate that our adapted images can be used to train a neural network to obtain much better performance on the target domain w.r.t the corresponding synthetic ones.  

Focusing on semantic segmentation, we have trained a standard FCN8s \cite{long2015fully} on the original GTA synthetic images and on our \textit{aligned} dataset. We tested both on the validation set of Cityscapes and reported the result in \autoref{tab:comparison}. For all our tests, we have initialized the feature extractor of the FCN8s with the publicly available VGG16 weights trained on the Imagenet dataset, then performed 100000 training iterations using batch 4, Adam optimizer and $0.0001$ as learning rate. We trained networks on 1024x1024 cropped images. 

To compare the networks we report two different metrics: the mean instance-level intersection-over-union (from now on shortened \textit{mIoU}) computed following the guidelines of the Cityscapes benchmark \cite{Cordts_2016_CVPR} and the overall pixel accuracy (shortened \textit{acc}), \ie, the percentage of correctly predicted pixel labels. We also report detailed scores for each semantic class  to highlight for which categories our image augmentation schema is more effective. We compare the results obtained by our domain adaptation method with alternatives recently proposed in literature: the feature-level alignment method of \cite{hoffman2016fcns}, the curriculum style domain adaptation approach of \cite{zhang2017curriculum} and the  pixel level alignment introduced in \cite{Hoffman_cycada2017}. In \autoref{tab:comparison} for all methods we report the performance achieved by training the very same FCN8s network \cite{long2015fully} both before  and after domain alignment, the former marked using \textit{Source} in the method column. For each row we report per class \textit{mIoU} and aggregated performance across the whole dataset (last two columns). Concerning aggregated \textit{mIoU} score, we can see how our proposal can outperform both \cite{zhang2017curriculum,hoffman2016fcns} while being comparable with \cite{Hoffman_cycada2017}.  Moreover, considering pixel accuracy, our proposal compares favourably even to \cite{Hoffman_cycada2017}. Considering the performance achieved before and after domain adaptation, our proposed pixel level alignment can provide an impressive $+16.9$ gain in \textit{mIoU} and a $+24$ in \textit{Acc.}, that, once again, compares favourably to \cite{hoffman2016fcns,zhang2017curriculum} and is  comparable to \cite{Hoffman_cycada2017}. Looking at class scores, we observe  how our proposal can achieve the best absolute performances on 10 classes out of 19, including some key ones for autonomous driving like \textit{road} ($+42.1$ gain between before and after alignment), car ($+46.4$) and \textit{person} ($+21.4$). We still lose something compared to other proposals on less common classes (\eg, \textit{bus}, \textit{motorcycle} and \textit{bicycle}), we think that this might be due to the dataset used not having enough samples of the target classes to effectively teach to the generator how to realistically render them. Even though our proposal performs comparably to \cite{Hoffman_cycada2017}, we would like to stress out how our adaptation method can be trained end-to-end instead of relying on separate training steps for the different parts.

In \autoref{fig:qualitative_segmentation} we also report some qualitative examples of the improvement in segmentation attainable by training on our \textit{adapted} GTA images (column (d)) compared to a purely synthetic training set (column (c)). Even if the results in column (d) are still far form optimal, most of the mistakes visible in column (c) are completely gone and the overall structure of the scene is more accurately segmented. Moreover, we can notice visually how the larger improvement concerns the segmentation of \textit{road} (colored purple), \textit{cars} (colored blue) and \textit{people} (colored red). 

\subsection{Ablation Study}
\label{sec:ablation}


\begin{table}
    \centering
    \scalebox{1.3}{
    \begin{tabular}{|l|cc|}
        \hline
        Test & mIoU & Acc.\\  
        \hline
        (a) Synthetic & 18.23 & 60.43 \\
        (b) GAN+Sem.  & 29.45 & 78.13 \\
        (c) GAN+Sem+weight. & 31.33 & 79.85 \\
        (d) Cycle \cite{Zhu_2017_ICCV} & 29.43 & 79.20 \\
        (e) Cycle+sem+weight. & \textbf{34.27} & \textbf{84.48} \\
        \hline
    \end{tabular}
    }
    \captionsetup{size=small,skip=0.333\baselineskip}
    \small
    \caption{Ablation study on the different component of our semantic aware GAN. Best results in bold.}
    \label{tab:ablation}
\end{table}

In \autoref{sec:segmentation_result} we have proven that the images generated by our proposal can effectively be used to train a semantic segmentation network so as to nearly double its performance compared to using  synthetic data only. In this section, instead, we  investigate more in depth on how each component of our proposal contributes to the final result.
Purposely,  we trained different architectures, keeping the comparison as fair as possible by maintaining the same training protocol. We report the results of these tests in \autoref{tab:ablation}.
We first investigated the performance of training a semantic segmentation network on images adapted by a simple GAN\cite{goodfellow2014generative}. As we obtained results even worse than our baseline network (a), we decided to not report them in \autoref{tab:ablation}. We then trained a GAN framework enriched with our semantic discriminator. Comparing line (b) with (a) we can clearly see how adding our semantics-aware discriminator not only allows to successfully train the GAN system but also results in a $+11.22\%$ \textit{mIoU}, thus testifying how semantic information can successfully regularize training. We then added our weighted L1 reconstruction loss between source and adapted image (c) slightly improving performances  by a $+1.88\%$ \textit{mIoU}. We then trained a standard CycleGAN \cite{Zhu_2017_ICCV} with no semantic clue demonstrating how having two couples of generator and discriminator is extremely effective to stabilize training of a GAN framework, as shown by (d) reaching comparable results to (b). Finally (e) reports the performance achievable by our full proposal that deploys the CycleGAN network combined with the semantics-aware discriminator and our semantic weighting system, achieving remarkable performance: $+16.04\%$ \textit{mIoU} and $+24.05\%$ \textit{Acc.} with respect to our baseline(a).

\section{Conclusion and Future Works}
In this paper we have demonstrated how a semantically aware image-to-image translation network can be successfully deployed to shrink the gap between images belonging to two drastically different domains such as synthetic and real images. Our novel network structure and loss function can successfully produce realistic images, thanks to its adversarial component, while at the same time maintaining structural coherence between input and output thanks to the enforced semantic consistency. 

We have addressed the adaptation problem only at \textit{pixel level}, however recent works \cite{Hoffman_cycada2017} have shown how for the semantic segmentation task the best absolute performance can be achieved by a mix of \textit{pixel level} and \textit{feature level} alignment. Therefore we plan to add an additional fine tuning step of our semantic segmentation network in order to introduce feature alignment in our pipeline. Moreover, we have tested our proposal for domain adaptation from synthetic to real images in the context of image segmentation, however, the same process can be used to address different tasks, \eg{} object detection, or different type of domain shifts, \eg{} different seasons, different sensors or different weather conditions. We plan to carry out these tests in order to  achieve a more comprehensive experimental evaluation of our proposal.  

{\small
    \bibliographystyle{IEEEtran}
    \bibliography{references}
}

\end{document}